\documentclass[letterpaper, 10 pt, conference]{ieeeconf}  

\usepackage[utf8]{inputenc}
\usepackage[T1]{fontenc}
\usepackage{graphicx}
\usepackage{hyperref}
\usepackage{amsmath} 
\usepackage{amssymb} 

\title{\LARGE \bf
Research Experiment on Multi-Model Comparison for Chinese Text Classification Tasks
}

\author{Li JiaCheng$^{1}$
}

\begin{document}

\maketitle
\thispagestyle{empty}
\pagestyle{empty}

\begin{abstract}

With the explosive growth of Chinese text data and advancements in natural language processing technologies, Chinese text classification has become one of the key techniques in fields such as information retrieval and sentiment analysis, attracting increasing attention. This paper conducts a comparative study on three deep learning models—TextCNN, TextRNN, and FastText—specifically for Chinese text classification tasks. By conducting experiments on the THUCNews dataset, the performance of these models is evaluated, and their applicability in different scenarios is discussed.

\end{abstract}

\section{INTRODUCTION}

In recent years, with the widespread adoption of the internet and mobile devices, effectively extracting valuable content from massive amounts of information has become a key concern for both academia and industry. Large volumes of unstructured Chinese text data, such as news reports and social media comments, are continuously emerging. As a fundamental task, Chinese text classification plays a vital role in improving information retrieval efficiency and supporting decision-making processes. This study focuses on the problem of Chinese text classification, aiming to explore the performance differences of various deep learning models in this task, thereby providing practical insights for real-world applications. Three representative models—TextCNN, TextRNN, and FastText—are selected for comparative experiments to facilitate understanding of their underlying mechanisms and technical characteristics.

\section{Related Technologies}
With the advancement of natural language processing (NLP) technologies, text classification has become a core task in fields such as information extraction, sentiment analysis, and recommendation systems. Due to the unique linguistic characteristics and technical requirements of the Chinese language, Chinese text classification has garnered extensive attention from researchers worldwide. In this domain, both academia and industry primarily focus on model optimization and practical applications.
\subsection{Classical Models' Application and Optimization}

Early text classification techniques were primarily based on statistical learning, with representative methods including Naive Bayes, Support Vector Machine (SVM), and Logistic Regression. These methods rely on manually constructed features (such as TF-IDF) for classification. While they perform well on small-scale datasets, they struggle to handle large volumes of data and complex contexts.

\subsection{Application of Deep Learning Models}

With the development of deep learning, models such as Convolutional Neural Networks (CNN), Recurrent Neural Networks (RNN), and Transformer-based architectures have been widely applied in text classification tasks, demonstrating significant advantages over traditional methods. 

\subsection{Application of Pre-trained Models}

With the introduction of the Transformer architecture and the widespread use of pre-training techniques, the performance of text classification has achieved a qualitative leap. For Chinese text classification tasks, researchers have developed pre-trained models that are tailored to the linguistic characteristics of the Chinese language, such as ERNIE, MacBERT, and PCL-Mixer. These models incorporate innovative technologies such as knowledge graphs and character-level representations, achieving excellent results in Chinese text classification tasks.

\section{ Model Design}

In this experiment, three representative models were selected for testing: TextCNN, TextRNN, and FastText. Below are the descriptions of each model.

\subsection{Text CNN} 

TextCNN (Text Convolutional Neural Network) is a deep learning model applied to text classification tasks, based on Convolutional Neural Networks (CNN) for processing text data. Compared to traditional natural language processing methods, TextCNN does not require manually designed complex feature engineering and can automatically extract features directly from the raw text.

\textbf{Here are the structure of model:}
\begin{itemize}

\item \textbf{Embedding Layer:}The first layer is a sentence matrix of size \(7 \times 57 \times 57 \times 5\), where each row represents a word's word vector. The dimension of each word vector is 5.
\item \textbf{Convolution Layer:}The model applies a one-dimensional convolutional layer with multiple convolutional kernels of varying sizes:
\begin{itemize}
    \item Convolutional kernel sizes: 2, 3, 4
    \item For each convolutional kernel size, the output channel number (\textit{out\_channel}) is set to 2.
\end{itemize}
\item \textbf{Max-Pooling Layer:}The third layer is a 1-max pooling layer. This operation ensures that sentences of different lengths can obtain a fixed-length feature representation. These fixed-length features are then concatenated together.
\item \textbf{Fully Connected and Softmax Layer:}Finally, the concatenated features are passed through a fully connected layer followed by the application of the Softmax function. The output is the probability distribution for each class.
\end{itemize}

\begin{figure}[ht!] 
    \centering
    \includegraphics[width=3.4in]{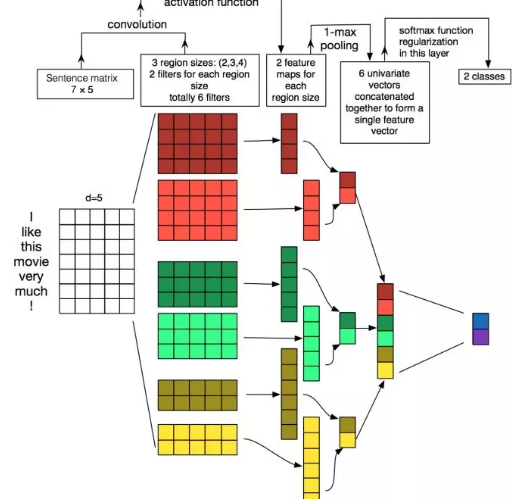}
    \caption{The structure of TextCNN}
    \label{bench}
\end{figure}


\subsection{TextRNN}
TextRNN (Text Recurrent Neural Network) is a deep learning model commonly used for text classification and other natural language processing tasks. Based on Recurrent Neural Networks (RNN), TextRNN effectively captures sequential information in text, making it particularly adept at handling contextual understanding and long-distance dependencies.

\textbf{Here are the structure of model:}
\begin{itemize}
    \item \textbf{Embedding Layer}:  
    Pre-trained word embeddings or randomly initialized embeddings are loaded.  
    \begin{itemize}
        \item Word embedding dimension:embed\_size
        \item Output:\([batch\_size, seq\_len, embed\_size]\)
    \end{itemize}
    
    \item \textbf{Bi-LSTM Layer}:  
    A bidirectional LSTM (Bi-LSTM) is used, with the hidden layer size set to \texttt{hidden\_size}.  
    \begin{itemize}
        \item All hidden states from both forward and backward passes are concatenated.
        \item Output:\([batch\_size,seq\_len, hidden\_size*2]\)
    \end{itemize}
    
    \item \textbf{Feature Concatenation (Concat Output)}:  
    The features are concatenated, and the number of prediction classes is set to num\_class.  
    \begin{itemize}
        \item Output:\([batch\_size,num\_class]\)
    \end{itemize}
    
    \item \textbf{Softmax Layer}:  
    Softmax normalization is applied, and the class with the maximum predicted value is selected as the final output.  
    \begin{itemize}
        \item Output:\([batch\_size, 1 ]\)
    \end{itemize}

    \begin{figure}[ht!] 
        \centering
        \includegraphics[width=3.4in]{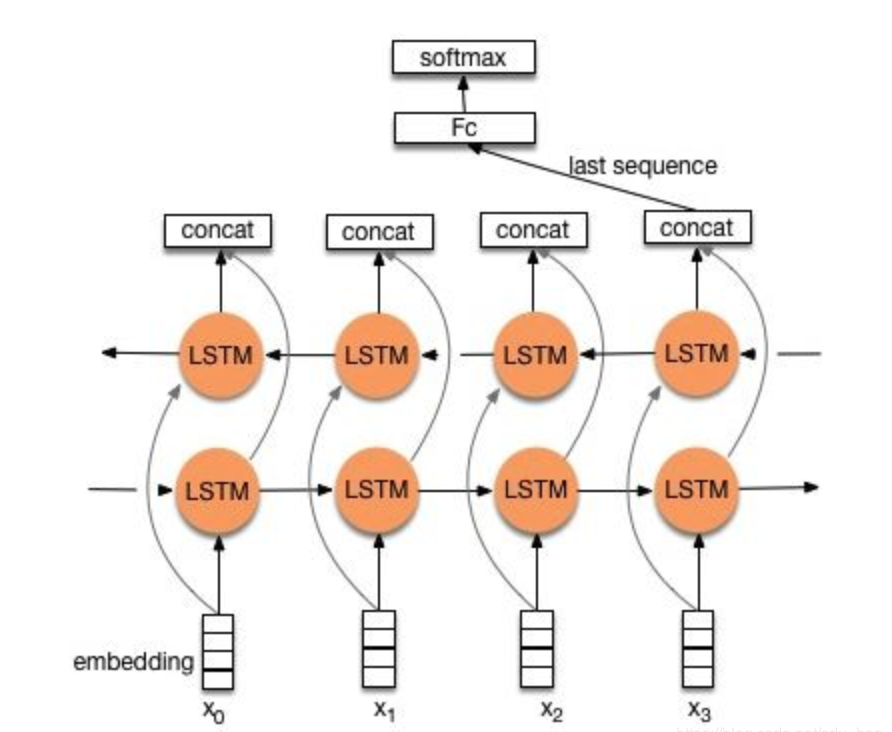}
        \caption{The structure of TextRNN}
        \label{bench}
    \end{figure}

\end{itemize}

\subsection{FastTEXT}
FastText is a library developed by Facebook AI Research for efficiently learning word vectors and performing text classification. It not only generates high-quality word embeddings but also supports simple text classification tasks. The design goal of FastText is to enable fast training and inference while maintaining good performance.

\textbf{Here are the structure of model:}
\begin{itemize}
    \item \textbf{Embedding Layer}:  
    Similar to many other models, FastText uses an embedding layer to map words into a continuous vector space. However, its uniqueness lies in its ability to handle Out-Of-Vocabulary (OOV) words, which are words not encountered during training.
    
    \item \textbf{n-gram Features}:  
    To capture word order information, FastText considers not only the representation of entire words but also breaks each word into a series of character-level n-grams.  
    \begin{itemize}
        \item For example, for the word "fast," if \( n = 3 \), the n-grams "fas" and "ast" are generated. 
        \item These n-grams are treated as independent features and have their own embeddings.
    \end{itemize}
    This approach allows the model to understand relationships between similarly spelled words and provides robustness against spelling errors.
    
    \item \textbf{Linear Classifier}:  
    FastText employs a simple linear classifier for text classification. Specifically:
    \begin{itemize}
        \item The embeddings of all words (including n-grams) in a document are summed or averaged to obtain the final document representation.
        \item This representation is then passed through a linear transformation followed by a softmax activation function to output the class probabilities.
    \end{itemize}
\end{itemize}

\begin{figure}[ht!] 
    \centering
    \includegraphics[width=3.4in]{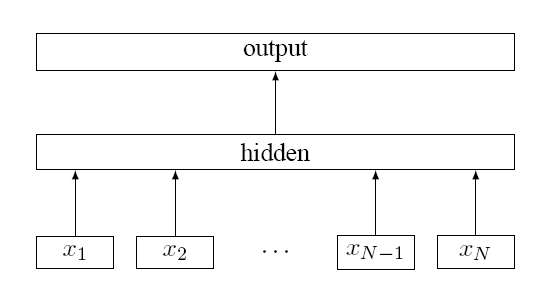}
    \caption{The structure of TextRNN}
    \label{bench}
\end{figure}

\section{The progress of experiment}
\subsection{Experimental Dataset}
The dataset used in this experiment originates from \textbf{THUCTC}, specifically the THUCNews news dataset. This dataset is based on historical data collected from the RSS subscription channels of Sina News between 2005 and 2011. After filtering and cleaning, it consists of approximately 740,000 news articles in UTF-8 encoded plain text format.

For this experiment, 200,000 news headlines were extracted, each with a length between 20 and 30 characters. These data were categorized into 10 classes, with 20,000 headlines per class. The specific categories include:
Finance,Real Estate,Stocks,Education,Technology,Society,Current Affairs,Sports,Gaming,Entertainment

\subsection{Dataset Extraction and Preprocessing}
The functionality for constructing the vocabulary, loading the dataset, data preprocessing, and building the data iterator is implemented through the untils.py and untilsfasttext.py files. Specifically, for the FastText model, the N-gram feature generation method is used for processing.

\subsection{Model Training}

The traineval.py file is used to call different models for training, complete the weight initialization, and perform function training and testing evaluation.

\section{Experimental Results and Analysis}
This experiment employs Cross-Entropy Loss, Accuracy, Precision, Recall, and F1-Score as evaluation metrics for analyzing the results. The detailed explanations of these metrics are as follows:

\subsection{Cross-Entropy Loss}
Cross-Entropy Loss measures the difference between the predicted probability distribution and the actual labels, making it particularly suitable for multi-class classification tasks. It is defined as:  
\[
Categorical Cross-Entropy = -\frac{1}{N} \sum_{i=1}^{N} \sum_{c=1}^{C} y_{ic} \log(\hat{y}_{ic})
\]
where \(N\) represents the total number of samples, \(C\) is the total number of classes, \(y_{ic}\) denotes the actual label, and \(\hat{y}_{ic}\) is the predicted probability.

\subsection{Accuracy}
Accuracy measures the proportion of correctly classified samples among the total number of samples. It is calculated as:  
\[
\mathrm{Accuracy} = \frac{\text{Number of Correctly Classified Samples}}{\text{Total Number of Samples}}
\]

\subsection{Precision}
Precision evaluates the proportion of correctly predicted samples for a specific class among all samples predicted to belong to that class. The formula is:  
\[
\mathrm{Precision} = \frac{\mathrm{True\ Positives\ (TP)}}{\mathrm{True\ Positives\ (TP)} + \mathrm{False\ Positives\ (FP)}}
\]
where:
\begin{itemize}
    \item \textbf{True Positives (TP):} The number of samples correctly predicted as belonging to the target class.
    \item \textbf{False Positives (FP):} The number of samples incorrectly predicted as belonging to the target class but actually belonging to other classes.
\end{itemize}

\subsection{Recall}
Recall measures the proportion of correctly predicted samples for a specific class among all actual samples of that class. The formula is:  
\[
\mathrm{Recall} = \frac{\mathrm{True\ Positives\ (TP)}}{\mathrm{True\ Positives\ (TP)} + \mathrm{False\ Negatives\ (FN)}}
\]
where:
\begin{itemize}
    \item \textbf{True Positives (TP):} The number of samples correctly predicted as belonging to the target class.
    \item \textbf{False Negatives (FN):} The number of samples that actually belong to the target class but are incorrectly predicted as belonging to other classes.
\end{itemize}

\subsection{F1-Score}
F1-Score is a comprehensive evaluation metric that combines Precision and Recall, providing a balanced assessment of the model's performance in terms of both accuracy and completeness. It is especially useful in scenarios where both precision and recall are equally important. The formula is:  
\[
\mathrm{F1-Score} = 2 \times \frac{\mathrm{Precision} \times \mathrm{Recall}}{\mathrm{Precision} + \mathrm{Recall}}
\]

\section{Results Display}
\subsection{Comparison of Test Loss and Test Accuracy Across Three Models}
After comparing the test loss and test accuracy of the three models, it was found that FastText achieved the best performance with a test accuracy of 92.02\%. This suggests that FastText outperforms the other models in handling the Chinese news headline text classification task. Both TextCNN and TextRNN performed similarly on these two metrics, with a test loss of 0.31 for both models, but TextCNN showed slightly better test accuracy.

The good performance of FastText can likely be attributed to its simple yet effective bag-of-words model combined with the use of n-gram features. This design allows FastText to capture local semantic information while maintaining high computational efficiency. In contrast, although TextCNN and TextRNN can also extract useful features, they did not surpass FastText in this particular experiment setup.
\begin{table}[h]
    \caption{Loss and Accuracy}
    \label{table_example}
    \begin{center}
    \begin{tabular}{|c||c||c|}
    \hline
    model name & Loss & Accuracy\\
    \hline
    TextCNN & 0.31 & 90.56\%\\
    \hline
    TextRNN & 0.31 & 90.41\%\\
    \hline
    FastText & 0.26 &92.02\%\\
    \hline
    \end{tabular}
    \end{center}
    \end{table}
\subsection{Comparison of Precision, Recall, and F1-Score Across Three Models}
In terms of overall performance, FastText achieved the highest accuracy across all categories, with a score of 92.02\%, followed by TextCNN at 90.56\%, and TextRNN at 90.41\%.
\begin{table}[h]
    \centering
    \caption{TextCNN}
    \label{tab:performance}
    \begin{tabular}{|c|c|c|c|}
    \hline
    Category       & Precision & Recall & F1-Score \\ \hline
    Finance        & 0.9368    & 0.8600 & 0.8968    \\ \hline
    Realty         & 0.9028    & 0.9380 & 0.9201    \\ \hline
    Stocks         & 0.8372    & 0.8640 & 0.8504    \\ \hline
    Education      & 0.9615    & 0.9480 & 0.9547    \\ \hline
    Science        & 0.8621    & 0.8630 & 0.8626    \\ \hline
    Society        & 0.8780    & 0.9280 & 0.9023    \\ \hline
    Politics       & 0.9094    & 0.8730 & 0.8908    \\ \hline
    Sports         & 0.9527    & 0.9460 & 0.9493    \\ \hline
    Game           & 0.9042    & 0.9160 & 0.9101    \\ \hline
    Entertainment & 0.9191    & 0.9200 & 0.9195    \\ \hline
    Overall Accuracy & 0.9056 & -      & -       \\ \hline
    \end{tabular}
    \end{table}
    
    \begin{table}[h]
        \centering
        \caption{TextRNN}
        \label{tab:performance}
        \begin{tabular}{|c|c|c|c|}
        \hline
        Category       & Precision & Recall & F1-Score \\ \hline
        Finance        & 0.9265    & 0.8570 & 0.8904    \\ \hline
        Realty         & 0.9132    & 0.9260 & 0.9196    \\ \hline
        Stocks         & 0.8636    & 0.8170 & 0.8397    \\ \hline
        Education      & 0.9405    & 0.9480 & 0.9442    \\ \hline
        Science        & 0.8544    & 0.8390 & 0.8466    \\ \hline
        Society        & 0.8752    & 0.9260 & 0.8999    \\ \hline
        Politics       & 0.8634    & 0.8910 & 0.8770    \\ \hline
        Sports         & 0.9866    & 0.9550 & 0.9705    \\ \hline
        Game           & 0.9234    & 0.9400 & 0.9316    \\ \hline
        Entertainment & 0.8980    & 0.9420 & 0.9195    \\ \hline
        Overall Accuracy & 0.9041 & -      & -       \\ \hline
        \end{tabular}
        \end{table}
        
        \begin{table}[h]
            \centering
            \caption{FastText}
            \label{tab:performance}
            \begin{tabular}{|c|c|c|c|}
            \hline
            Category       & Precision & Recall & F1-Score \\ \hline
            Finance        & 0.9303    & 0.8810 & 0.9050    \\ \hline
            Realty         & 0.9412    & 0.9280 & 0.9345    \\ \hline
            Stocks         & 0.8459    & 0.8780 & 0.8616    \\ \hline
            Education      & 0.9464    & 0.9540 & 0.9502    \\ \hline
            Science        & 0.8741    & 0.8820 & 0.8780    \\ \hline
            Society        & 0.9023    & 0.9240 & 0.9130    \\ \hline
            Politics       & 0.8919    & 0.9080 & 0.8999    \\ \hline
            Sports         & 0.9827    & 0.9660 & 0.9743    \\ \hline
            Game           & 0.9544    & 0.9420 & 0.9482    \\ \hline
            Entertainment & 0.9390    & 0.9390 & 0.9390    \\ \hline
            Overall Accuracy & 0.9202 & -      & -       \\ \hline
            \end{tabular}
            \end{table}
    
\subsection{Experimental Results Analysis}       
Overall, FastText demonstrates the best performance in terms of overall accuracy and the majority of categories for precision, recall, and F1-score, particularly excelling in categories like **realty**, **education**, and **politics**. This advantage is attributed to its simple yet efficient bag-of-words model combined with n-gram features, which enables it to capture critical local semantic information while maintaining computational efficiency. 

TextCNN also performs commendably in most categories, especially in **stocks** and **society**, where it shows results comparable to FastText. However, it slightly lags behind in overall accuracy. 

Although TextRNN is somewhat outperformed by the other two models overall, it exhibits exceptional performance in categories like **sports**, which could be due to its ability to effectively capture sequential information and long-term dependencies.



\begin{thebibliography}{10}

    \bibitem{chen2023}
    Chong Chen, Zijia Cheng, Chuanqing Wang, and Lei Li.
    \newblock {Key Point Identification and Utilization in Scientific Papers Based on Peer Review Comments}.
    \newblock {\em Journal of Information Science}, 2023.
    
    \bibitem{fusion2023}
    Authors Unknown.
    \newblock {A Multi-Modal Emotion Recognition Method Combining Feature and Decision Fusion}.
    
    \bibitem{shilei2021}
    Lei Shi, Mingyu Wang, Zheli Song, Yongcai Tao, Lin Wei, Yufei Gao, and Yuxin Fan.
    \newblock {Text Classification Research Combining Self-Attention Mechanism and BiGRU}.
    \newblock {\em Mini-Micro Computer Systems}, 2021.
    
    \bibitem{wang2022}
    Li'e Wang, Xiaocong Li, and Hongyi Liu.
    \newblock {News Recommendation Method Integrating Knowledge Graphs and Differential Privacy}.
    \newblock {\em Computer Applications}, 2022.
    
    \bibitem{csdn_machinelearning}
    BelthPixtink.
    \newblock {Machine Learning Models}.

    
    \bibitem{capsule_bigru}
    Qi Zhao, Yanhui Du, Tianliang Lu, and Shaoyu Shen.
    \newblock {Text Similarity Analysis Algorithm Based on Capsule-BiGRU}.
    \newblock {\em Computer Engineering and Applications}, 2021.
    
    \bibitem{fasttext_csdn}
    Sonhhxg.
    \newblock {Fast Text Classification (FastText)}.

    
    \bibitem{bert_deepconv}
    Wenhao Yang, Guangcong Liu, and Kejin Luo.
    \newblock {News Label Classification Based on BERT and Deep Convolution}.
    \newblock {\em Computer and Modernization}, 2021.
    
\end{thebibliography}
\end{document}